\RequirePackage{fix-cm}
\documentclass[twocolumn]{svjour3}          
\smartqed  

\usepackage{cite}
\usepackage[pdftex]{graphicx}
\usepackage{ragged2e}
\usepackage[tight,footnotesize]{subfigure}
\usepackage{rotating}
\usepackage{graphicx}
\usepackage{amsmath,amssymb} 
\usepackage{array}
\usepackage{multirow}
\usepackage{colortbl}
\usepackage{amsfonts}
\usepackage{pifont}
\usepackage{xspace}
\usepackage{etoolbox}
\usepackage{overpic}
\usepackage{color}
\usepackage{microtype}

\usepackage{graphicx}%
\usepackage{multirow}%
\usepackage{amsmath,amssymb,amsfonts}%
\usepackage{mathrsfs}%
\usepackage[title]{appendix}%
\usepackage{xcolor}%
\usepackage{textcomp}%
\usepackage{manyfoot}%
\usepackage{booktabs}%
\usepackage{algorithm}%
\usepackage{algorithmicx}%
\usepackage{algpseudocode}%
\usepackage{listings}%
\usepackage{multirow}
\usepackage{makecell}
\usepackage{graphicx}
\usepackage{xspace}
\usepackage{xcolor}
\usepackage{amssymb}
\usepackage{color}
\usepackage{graphicx}
\usepackage{booktabs}

\newcommand{\orcid}[1]{\href{https://orcid.org/#1}{\includegraphics[width=10pt]{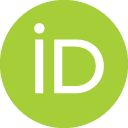}}}

\def\etal{{\em et al}}

\usepackage{hyperref}
\hypersetup{breaklinks=true,citecolor=blue, colorlinks}

\DeclareGraphicsExtensions{.pdf,.jpg,.png}

\usepackage{silence}
\journalname{Research Article}

\begin{document}

\title{An Empirical Study of LLaMA3 Quantization: From LLMs to MLLMs}

\titlerunning{An Empirical Study of LLaMA3 Quantization: From LLMs to MLLMs}        











\author{
{Wei} {Huang}$^\dag$ \orcid{0009-0007-9885-0028}\and
{Xingyu} {Zheng}$^\dag$ \orcid{0009-0009-6283-7635} \and 
{Xudong} {Ma}$^\dag$ \orcid{0000-0002-7985-3350} \and 
{Haotong} {Qin}$^*$ \orcid{0000-0001-7391-7539} \and 
{Chengtao} {Lv} \orcid{0000-0001-9599-6557} \and 
{Hong} {Chen} \orcid{0009-0001-9658-0593} \and 
{Jie} {Luo} \orcid{0000-0002-4157-9931} \and 
{Xiaojuan} {Qi} \orcid{0000-0002-4285-1626} \and 
{Xianglong} {Liu} \orcid{0000-0002-7618-3275} \and 
{Michele} {Magno} \orcid{0000-0003-0368-8923} \and 
}

\authorrunning{F. Author \etal} 

\institute{
{Wei} {Huang} and {Xiaojuan} {Qi} are with institution, Department of Electrical and Electronic Engineering, The University of Hong Kong, Pokfulam Road, Hong Kong, 999077, China.
(Email: weih@connect.hku.hk, xjqi@eee.hku.hk). \\
{Xingyu} {Zheng}, {Xudong} {Ma}, {Chengtao} {Lv}, {Hong} {Chen}, {Jie} {Luo} and {Xianglong} {Liu} are with institution, School of Computer Science and Engineering, Beihang University, Xueyuan Road, Beijing, 100191, China.
(Email: xingyuzheng@buaa.edu.cn, macaronlin@buaa.edu.cn, lvchengtao@buaa.edu.cn, 18373205@buaa.edu.cn, luojie@buaa.edu.cn, xlliu@buaa.edu.cn). \\
{Haotong} {Qin} and {Michele} {Magno} are with institution, Department of Information Technology and Electrical Engineering, ETH Zurich, Sternwartstrasse 7, Z¨urich, Switzerland.
(Email: haotong.qin@pbl.ee.ethz.ch, michele.magno@pbl.ee.ethz.ch). \\
$^*$ Corresponding author: Haotong Qin. \\
$^\dag$ These authors contributed equally to this work.
}

\date{Received: date / Accepted: date}

\maketitle

\begin{abstract}
The LLaMA family, a collection of foundation language models ranging from 7B to 65B parameters, has become one of the most powerful open-source large language models (LLMs) and the popular LLM backbone of multi-modal large language models (MLLMs), widely used in computer vision and natural language understanding tasks. In particular, LLaMA3 models have recently been released and have achieved impressive performance in various domains with super-large scale pre-training on over 15T tokens of data. Given the wide application of low-bit quantization for LLMs in resource-constrained scenarios, we explore LLaMA3's capabilities when quantized to low bit-width. This exploration can potentially provide new insights and challenges for the low-bit quantization of LLaMA3 and other future LLMs, especially in addressing performance degradation issues that suffer in LLM compression. Specifically, we comprehensively evaluate the 10 existing post-training quantization and LoRA fine-tuning (LoRA-FT) methods of LLaMA3 on 1-8 bits and various datasets to reveal the low-bit quantization performance of LLaMA3. To uncover the capabilities of low-bit quantized MLLM, we assessed the performance of the LLaMA3-based LLaVA-Next-8B model under 2-4 ultra-low bits with post-training quantization methods. Our experimental results indicate that LLaMA3 still suffers from non-negligible degradation in linguistic and visual contexts, particularly under ultra-low bit widths. This highlights the significant performance gap at low bit-width that needs to be addressed in future developments. We expect that this empirical study will prove valuable in advancing future models, driving LLMs and MLLMs to achieve higher accuracy at lower bit to enhance practicality. Our project is released on \href{https://github.com/Macaronlin/LLaMA3-Quantization}{GitHub} and quantized models are released in \href{https://huggingface.co/Efficient-ML}{HuggingFace}.


\end{abstract}

\section{Introduction}
\label{sec1}

Launched by Meta in February 2023, the LLaMA~\cite{touvron2023llama} series\footnote{\ \url{https://llama.meta.com}\label{foot:llama}}, a collection of foundation language models ranging from 7B to 65B parameters, represents a breakthrough in autoregressive large language models (LLMs) using the Transformer~\cite{vaswani2017attention} architecture. From its first release, with 13 billion parameters, it outperformed the much larger, closed-source GPT-3 model with 175 billion parameters. On April 18, 2024, Meta introduced the LLaMA3 model, offering 8 billion and 70 billion parameter configurations. Thanks to extensive pre-training on more than 15 trillion data tokens, the LLaMA3 models~\cite{dubey2024llama} have achieved state-of-the-art performance across a wide range of tasks, establishing the LLaMA family as one of the best open-source LLMs available for a wide variety of applications and deployment scenarios. Recently, the LLaVA team~\cite{liu2023llava} has launched the new LLaVA-Next-8B\footnote{\ \url{https://llava-vl.github.io/blog/2024-05-10-llava-next-stronger-llms}\label{foot:llava-next}} model based on LLaMA3, giving the stronger general multi-modal capabilities of multi-modal large language models (MLLMs).

Despite their impressive performance, deploying LLaMA3 models still poses significant challenges due to resource limitations in many scenarios. Fortunately, low-bit quantization~\cite{xiao2023smoothquant,qin2024quantsr,jacob2018quantization,huang2024slim} has emerged as one of the most popular techniques for compressing LLMs. This technique reduces the memory and computational requirements of LLMs during inference, enabling them to run on resource-limited devices. Addressing the performance drop after compression is a major concern for current LLM quantization approaches. While numerous low-bit quantization methods have been proposed, their evaluations have primarily focused on the earlier and less capable LLaMA models (LLaMA and LLaMA2). Thus, LLaMA3 presents a new opportunity for the LLM community to assess the performance of quantization on cutting-edge LLMs and MLLMs and understand existing methods' strengths and limitations. In this empirical study, we aim to analyze the capability of LLaMA3 to handle the challenges associated with degradation due to quantization. 

\begin{figure*}[t]
  \centering
  \includegraphics[width=\textwidth]{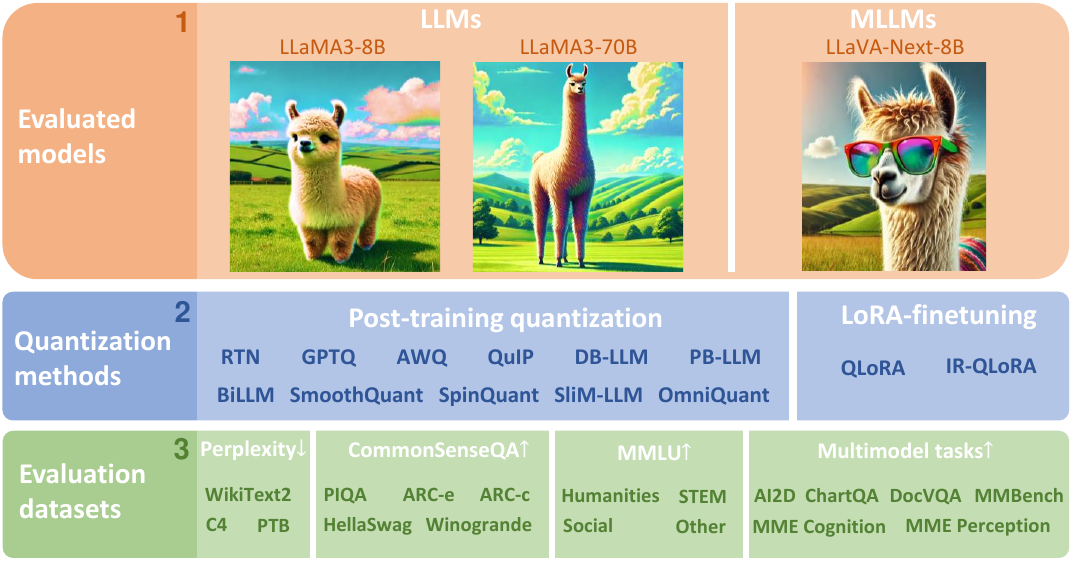}
  \caption{The overview of our empirical study}
  \label{fig:overview}
\end{figure*}

Our study delineates the outcomes of two principal techniques for quantizing LLaMA3 across three evaluation tracks: post-training quantization (PTQ) of LLMs, quantization of LLMs via LoRA-FineTuning (LoRA-FT), and PTQ of LLaMA3-based MLLM, aiming to conduct a comprehensive assessment of the LLaMA3 model's capabilities in language and visual-language tasks. We explore a range of cutting-edge quantization methods across technical tracks (RTN~\cite{nagel2020up}, GPTQ~\cite{frantar2022gptq}, AWQ~\cite{lin2023awq}, SmoothQuant~\cite{xiao2023smoothquant}, PB-LLM~\cite{shang2023pb}, QuIP~\cite{chee2024quip}, DB-LLM~\cite{chen2024db}, BiLLM~\cite{huang2024billm}, and SliM-LLM~\cite{huang2024slim} for PTQ; QLoRA~\cite{dettmers2024qlora} and IR-QLoRA~\cite{qin2024accurate} for LoRA-FT), covering a wide spectrum from 1 to 8 bits and utilizing a diverse array of evaluation datasets, including WikiText2~\cite{merity2016pointer}, C4~\cite{raffel2020exploring}, PTB~\cite{marcus1994penn}, CommonSenseQA datasets (PIQA~\cite{bisk2020piqa}, ARC-e~\cite{clark2018think}, ARC-c~\cite{clark2018think}, HellaSwag~\cite{zellers2019hellaswag}, Winogrande~\cite{sakaguchi2021winogrande}), and MMLU~\cite{hendryckstest2021} benchmark. For multi-modal tasks, we follow a common practice~\cite{lin2023awq}, performing low-bit post-training quantization on the LLM component of LLaVA-Next-8B using GPTQ and AWQ. We then validate the quantized MLLM inference capabilities on 6 visual language benchmarks, including AI2D~\cite{kembhavi2016diagram}, ChartQA~\cite{masry2022chartqa}, DocVQA~\cite{mathew2021docvqa}, MME~\cite{fu2024mmecomprehensiveevaluationbenchmark}, and MMBench(English)~\cite{liu2025mmbench}. These evaluations assess the capabilities and limitations of the LLaMA3 model under current LLM quantization techniques and serve as a source of inspiration for designing future large language and large visual-language model quantization methods.
The decision to focus specifically on the LLaMA3 model is motivated by its superior performance among all current open-source instruction-tuned LLMs on a variety of datasets, including 5-shot MMLU, 0-shot GPQA, 0-shot HumanEval, 8-shot CoT GSM-8K, and 4-shot CoT MATH. The overview of our study is presented as Fig.~\ref{fig:overview}.

This not only helps advance the research within the LLM and MLLM quantization community, but also facilitates a broader understanding and application of effective quantization.

We evaluate the low-bit quantization of LLaMA3-8B, -70B, and LLaVA-Next-8B, where the pre-trained models were obtained from their official repositories\textsuperscript{\ref{foot:llava-next}}.

\paragraph{Quantization methods\quad} 
To evaluate the performance of low-bit quantized LLaMA3, we select representative LLM quantization methods with extensive influence and functionality, including 9 PTQ methods and 2 LoRA-FT methods. The implementations of our evaluated quantization methods follow their open-source repositories\footnote{\ \url{https://github.com/IST-DASLab/gptq},\url{https://github.com/mit-han-lab/llm-awq},\url{https://github.com/mit-han-lab/smoothquant}, \url{https://github.com/Cornell-RelaxML/QuIP}, \url{https://github.com/Aaronhuang-778/SliM-LLM}, \url{https://github.com/hahnyuan/PB-LLM}, \url{https://github.com/Aaronhuang-778/BiLLM}, \url{https://github.com/artidoro/qlora}, \url{https://github.com/htqin/IR-QLoRA}}. We also used 8 NVIDIA A800 with 80GB GPU memory for quantitative evaluation.

\paragraph{Evaluation datasets\quad} 
For the PTQ methods, we evaluate quantized LLaMA3 on the WikiText2~\cite{merity2016pointer}, PTB~\cite{marcus1994penn}, and a portion of the C4 dataset~\cite{raffel2020exploring}, using perplexity (PPL) as the evaluation metric. Subsequently, we further conduct experiments on five zero-shot evaluation tasks (PIQA~\cite{bisk2020piqa}, Winogrande~\cite{sakaguchi2021winogrande}, ARC-e ~\cite{clark2018think}, ARC-c~\cite{clark2018think}, and Hellaswag~\cite{zellers2019hellaswag}) to fully validate the quantized performance of LLaMA3. We further conduct the evaluation on 5 visual language benchmarks (AI2D, ChartQA, DocVQA, MME, and MMBench(English)) for quantized LLaVA-Next-8B. {To ensure fairness in evaluation of different PTQ methods, we set WikiText2 as the calibration dataset for all quantization methods, with a sample size of 128 and a sequence length of 2048. Additionally, for methods requiring grouped quantization, we standardize the block size at 128 to balance performance and inference efficiency, a common practice in existing studies.} For the LoRA-FT methods, we conduct the evaluation on the 5-shot MMLU benchmark~\cite{hendrycks2020measuring} while also validating the aforementioned 5 zero-shot datasets for the LoRA-FT methods. {To ensure fairness in the evaluation of different LoRA-FT methods, we fine-tune all models using the same training data and consistent hyperparameters, including learning rate, batch size, number of training epochs, and LoRA configurations such as rank and scaling factors.}

\section{Track1: Post-Training Quantization}\label{sec:ptq}

\textbf{Quantization framework.} {We begin by outlining the general uniform quantization process for LLMs, following standard practices as described in Refs.~\cite{liu2023llm,frantar2022gptq,xiao2023smoothquant}. This process involves mapping floating-point weights, distributed within the range $[w_{\mathrm{min}}, w_{\mathrm{max}}]$, to an integer range of $2^N$, where $N$ denotes the target bit-width. The quantization function for a weight matrix $\boldsymbol{w}_{f} \in \mathbb{R}^{n \times m}$ is defined as follows:}

 \begin{subequations} \label{eq:quantization_ptq}
 \begin{align}
\hat{\boldsymbol{w}}_q &= \operatorname{clamp}(\lfloor \frac{\boldsymbol{w}_f }{\Delta} \rceil + z, 0, 2^N - 1)  \\
\Delta &= \frac{w_\mathrm{max} - w_\mathrm{min}}{2^N - 1} \\
z &= - \lfloor \frac{w_\mathrm{min}}{\Delta} \rceil 
\end{align}
 \end{subequations}

where $\hat{\boldsymbol{w}}_q$ indicates quantized weight, which is integer, $N$ denotes the target bit-width, $\lfloor \cdot \rceil$ is round operation and $\operatorname{clamp}(\cdot)$ constrains the value within integer range (e.g. $[0,1,2,3]$, $N=2$). $\Delta$ is scale factor and $z$ is quantization zero point, respectively. As shown in Table~\ref{tab:ptq_lm38_1} to Table~\ref{tab:ptq_lm370_2}, we provide the performance of the low-bit LLaMA3-8B and LLaMA3-70B with 8 different PTQ methods, respectively, covering a wide bit-width spectrum from 1 to 8 bits. In addition, the performance of LLaMA1 and LLaMA2 under the same setting are summarized in Table~\ref{tab:ptq_lm8_1}.

\begin{table*}[!t]
    \small
    \centering
    \caption{Evaluation results of post-training quantization on the LLaMA3-8B model (1/2). {\\ \#W, \#A, and \#G represent the bit-width for weight, activation, and group size, respectively, '-' indicates no grouping required, and $\downarrow$ denotes that the lower is better.}}

    \label{tab:ptq_lm38_1}
    \setlength{\tabcolsep}{6.8mm}
    \renewcommand\arraystretch{0.83}
    {
    \begin{tabular}{lllrrrr}
        \toprule
        \multirow{2}{*}{{Method}} & 
        \multirow{2}{*}{{\#W }} & 
        \multirow{2}{*}{{\#A}} & 
        \multirow{2}{*}{{\#G}}&
        \multicolumn{3}{c}{{PPL$\downarrow$}}
        \\
        \cmidrule(lr){5-7}
        ~ & ~ & ~ & ~ & {WikiText2} & {C4} & {PTB} \\ 
        \midrule
        LLaMA3 & 16 & 16 & - & 6.1 & 9.2 & 10.6 \\
        \midrule
        \midrule
        
        \multirow{7}{*}{RTN} 
         & 4 & 16 & 128 & 8.5 & 13.4 & 14.5 \\
         & 3 & 16 & 128 & 27.9 & 1.1e2 & 95.6 \\ 
         & 2 & 16 & 128 & 1.9\(\times 10^3\) & 2.5\(\times 10^4\) & 1.8\(\times 10^4\) \\
         & 8 & 16 & - & 6.2 & 9.5 & 11.2 \\
         & 4 & 16 & - & 8.7 & 14.0 & 14.9 \\
         & 3 & 16 & - & 2.2\(\times 10^3\) & 5.6\(\times 10^2\) & 2.0\(\times 10^3\) \\ 
         & 2 & 16 & - & 2.7\(\times 10^6\) & 7.4\(\times 10^6\) & 3.1\(\times 10^6\) \\
        \midrule
        
        \multirow{7}{*}{GPTQ~\cite{frantar2022gptq}} 
         &  4  & 16 & 128 & 6.5 & 10.4 & 11.0 \\
         &  3  & 16 & 128 & 8.2 & 13.7 & 15.2 \\
         &  2  & 16 & 128 & 2.1\(\times 10^2\) & 4.1\(\times 10^4\) & 9.1\(\times 10^2\) \\
         &  8  & 16 & - & 6.1 & 9.4 & 10.6 \\
         &  4  & 16 & - & 7.0 & 11.8 & 14.4 \\
         &  3  & 16 & - & 13.0 & 45.9 & 37.0 \\
         &  2  & 16 & - & 5.7\(\times 10^4\) & 1.0\(\times 10^5\) & 2.7\(\times 10^5\) \\
        \midrule

        \multirow{7}{*}{AWQ~\cite{lin2023awq}} 
        & 4 & 16 & 128 & 6.6 & 9.4 & 11.1 \\
        & 3 & 16 & 128 & 8.2 & 11.6 & 13.2 \\
        & 2 & 16 & 128 & 1.7\(\times 10^6\) & 2.1\(\times 10^6\) & 1.8\(\times 10^6\) \\
        & 8 & 16 & - & 6.1 & 8.9 & 10.6 \\
        & 4 & 16 & -  & 7.1 & 10.1 & 11.8 \\
        & 3 & 16 & - & 12.8 & 16.8 & 24.0 \\
        & 2 & 16 & - & 8.2\(\times 10^5\) & 8.1\(\times 10^5\) & 9.0\(\times 10^5\) \\
        \midrule

        \multirow{3}{*}{SliM-LLM~\cite{huang2024slim}} 
        & 4 & 16 & 128 & 6.4 & 9.5 & 10.9 \\
        & 3 & 16 & 128 & 7.7 & 13.1 & 14.7 \\
        & 2 & 16 & 128 & 39.7 & 1.1\(\times 10^2\) & 1.6\(\times 10^2\)  \\ 
        \midrule
        
        \multirow{3}{*}{QuIP~\cite{chee2024quip}} 
        & 4 & 16 & - & 6.5 & 11.1 & 9.5  \\
        & 3 & 16 & - & 7.5 & 11.3 & 12.6 \\
        & 2 & 16 & - & 85.1 & 1.3\(\times 10^2\) & 1.8\(\times 10^2\) \\ 
        \midrule
        
        DB-LLM~\cite{chen2024db} & 2 & 16 & 128 & 13.6 & 19.2 & 23.8 \\
        \midrule
        
        \multirow{2}{*}{PB-LLM~\cite{shang2023pb}}
        & 2 & 16 & 128 & 24.7 & 79.2 & 65.6 \\
        & 1.7 & 16 & 128 & 41.8 & 2.6\(\times 10^2\) & 1.2\(\times 10^2\) \\
        \midrule
        
        BiLLM~\cite{huang2024billm} & 1.1 & 16 & 128 & 28.3 & 2.9\(\times 10^2\) & 94.7 \\
        \midrule
        \midrule

        \multirow{3}{*}{SmoothQuant~\cite{xiao2023smoothquant}} 
        & 8 & 8 & - & 6.3 & 9.2 & 10.8 \\
        & 6 & 6 & - & 7.7 & 11.8 & 12.5 \\
        & 4 & 4 & - & 4.3\(\times 10^3\) & 4.0\(\times 10^3\) & 3.6\(\times 10^3\) \\
        \midrule 
        
        \multirow{2}{*}{OmniQuant~\cite{shao2023omniquant}} 
        &  6 &  6 &  - &  7.0 &  10.1 &  - \\
        &  4 &  4 &  - &  4.4\(\times 10^2\) &  3.2\(\times 10^2\) &  - \\ 
        \midrule 
        
        \multirow{2}{*}{I-LLM~\cite{hu2024llm}} 
        &  6 &  6 &  - &  6.6 &  9.8 &  - \\
        &  4 &  4 &  - &  21.2 &  30.9 &  - \\ 
        \midrule 
        
        \multirow{2}{*}{SpinQuant~\cite{liu2024spinquant}} 
        &  4 &  8 &  - &  6.5 &  - &  - \\
        &  4 &  4 &  - &  7.1 &  - &  - \\ 
        
        \bottomrule
        
    \end{tabular}}
\end{table*}

\begin{table*}[!t]
    \small
    \centering
    \caption{Evaluation results of post-training quantization on the LLaMA3-70B model (1/2)}

    \label{tab:ptq_lm370_1}
    \setlength{\tabcolsep}{6.8mm}
    {
    \begin{tabular}{lllrrrr}
        \toprule
        \multirow{2}{*}{{Method}} & 
        \multirow{2}{*}{{\#W }} & 
        \multirow{2}{*}{{\#A}} & 
        \multirow{2}{*}{{\#G}}&
        \multicolumn{3}{c}{{PPL$\downarrow$}}
        \\
        \cmidrule(lr){5-7}
        ~ & ~ & ~ & ~ & {WikiText2} & {C4} & {PTB} \\

        \midrule
        LLaMA3 & 16 & 16 & - & 2.9 & 6.9 & 8.2\\
        \midrule
        \midrule
        
        \multirow{3}{*}{RTN} 
        & 4 & 16 & 128 & 3.6 & 8.9 & 9.1 \\
        & 3 & 16 & 128 & 11.8 & 22.0 & 26.3 \\
        & 2 & 16 & 128 & 4.6\(\times 10^5\) & 4.7\(\times 10^5\) & 3.8\(\times 10^5\)  \\
        \midrule
        
        \multirow{3}{*}{GPTQ~\cite{frantar2022gptq}} 
        & 4 & 16 & 128 & 3.3 & 6.9 & 8.3 \\
        & 3 & 16 & 128 &  5.2&10.5  &9.7 \\
        & 2 & 16 & 128 & 11.9 & 22.8 & 31.6 \\
        \midrule

        \multirow{3}{*}{AWQ~\cite{lin2023awq}} 
        & 4 & 16 & 128 & 3.3 & 7.0 & 8.3 \\
        & 3 & 16 & 128 & 4.8 & 8.0 & 9.0 \\
        & 2 & 16 & 128 & 1.7\(\times 10^6\) & 1.4\(\times 10^6\) & 1.5\(\times 10^6\) \\
        \midrule

        \multirow{3}{*}{SliM-LLM~\cite{huang2024slim}} 
        & 4 & 16 &  128  & 3.3 & 7.0 & 8.3  \\
        & 3 & 16 &  128  & 4.1 & 7.9 & 9.0  \\
        & 2 & 16 &  128  & 9.5 & 16.2 & 18.7 \\ 
        \midrule
        
        \multirow{3}{*}{QuIP~\cite{chee2024quip}} 
        & 4 & 16 &  -  & 3.4 & 7.1 & 8.4  \\
        & 3 & 16 &  -  & 4.7 & 8.0 & 8.9  \\
        & 2 & 16 &  -  & 13.0 & 22.2 & 24.9  \\

        \midrule
        
        \multirow{2}{*}{PB-LLM~\cite{shang2023pb}}
        & 2 & 16 & 128 & 11.6 & 34.5 & 27.2  \\
        & 1.7 & 16 & 128 &  18.6 & 65.2 & 55.9 \\
        \midrule
        
        BiLLM~\cite{huang2024billm} & 1.1 & 16 & 128 & 17.1 & 77.7 & 54.2\\ 
        \midrule
        \midrule

        \multirow{3}{*}{SmoothQuant~\cite{xiao2023smoothquant}} 
        & 8 & 8 & - & 2.9 & 6.9 & 8.2 \\
        & 6 & 6 & - & 2.9 & 6.9 & 8.2 \\
        & 4 & 4 & - & 9.6 & 16.9 & 17.7 \\   
        
        \bottomrule
        
    \end{tabular}}
\end{table*}

\begin{table*}[!t]
    \small
    \centering
    \caption{Evaluation results of post-training quantization on LLaMA3-8B model (2/2). $\uparrow$ indicates that the higher value is better.}

    \label{tab:ptq_lm38_2}
    \setlength{\tabcolsep}{3.9mm}
    {
    \begin{tabular}{lllrrrrrrr}
        \toprule
        \multirow{2}{*}{{Method}} & 
        \multirow{2}{*}{{\#W }} & 
        \multirow{2}{*}{{\#A}} & 
        \multirow{2}{*}{{\#G}}&
        \multicolumn{6}{c}{{CommonSenseQA$\uparrow$}} 
        \\
        \cmidrule(lr){5-10}
        ~ & ~ & ~ & ~ & {PIQA} &{ARC-e} & {ARC-c} &{HellaSwag} & {Wino} & {Avg.} \\ 
        \midrule
        LLaMA3 & 16 & 16 & - & 79.9 & 80.1 & 50.4 & 60.2 & 72.8 & 68.6 \\
        \midrule
        \midrule
        
        \multirow{7}{*}{RTN} 
         & 4 & 16 & 128 
         & 76.6 & 70.1 & 45.0 & 56.8 & 71.0 & 63.9 \\
         & 3 & 16 & 128 
         & 62.3 & 32.1 & 22.5 & 29.1 & 54.7 & 40.2 \\
         & 2 & 16 & 128   
         & 53.1 & 24.8 & 22.1 & 26.9 & 53.1 & 36.0 \\
         & 8 & 16 & - 
         & 79.7 & 80.8 & 50.4 & 60.1 & 73.4 & 68.9 \\
         & 4 & 16 & -   
         & 75.0 & 68.2 & 39.4 & 56.0 & 69.0 & 61.5 \\
         & 3 & 16 & -  
         & 56.2 & 31.1 & 20.0 & 27.5 & 53.1 & 35.6 \\ 
         & 2 & 16 & -  
         & 53.1 & 24.7 & 21.9 & 25.6 & 51.1 & 35.3 \\
        \midrule
        
        \multirow{7}{*}{GPTQ~\cite{frantar2022gptq}} 
         &  4  &  16  &  128  
         & 78.4 & 78.8 & 47.7 & 59.0 & 72.6 & 67.3  \\
         &  3  &  16  &  128  
         & 74.9 & 70.5 & 37.7 & 54.3 & 71.1 & 61.7  \\
         &  2  &  16  &  128  
         & 53.9 & 28.8 & 19.9 & 27.7 & 50.5 & 36.2  \\
         &  8  &  16  &  -  
         & 79.8 & 80.1 & 50.2 & 60.2 & 72.8 & 68.6  \\
         &  4  &  16   &  -  
         &  76.8  &  74.3  &  42.4  &  57.4 &  72.8  &  64.8\\
         &  3  &  16  &  -    
         &  60.8  &  38.8  &  22.3  &  41.8  &  60.9  &  44.9\\
         &  2  &  16  &  -  
         & 52.8 & 25.0 & 20.5 & 26.6 & 49.6 & 34.9\\
        \midrule

        \multirow{7}{*}{AWQ~\cite{lin2023awq}} 
        & 4 & 16 & 128 & 79.1 & 79.7 & 49.3 & 59.1 & 74.0 & 68.2 \\
        & 3 & 16 & 128 & 77.7 & 74.0 & 43.2 & 55.1 & 72.1 & 64.4 \\
        & 2 & 16 & 128  & 52.4 & 24.2 & 21.5 & 25.6 & 50.7 & 34.9 \\
        & 8 & 16 &  -   & 79.6 & 80.3 & 50.5 & 60.2 & 72.8 & 68.7 \\
        & 4 & 16 &  -  & 78.3 & 77.6 & 48.3 & 58.6 & 72.5 & 67.0 \\
        & 3 & 16 &  -  & 71.9 & 66.7 & 35.1 & 50.7 & 64.7 & 57.8 \\
        & 2 & 16 &  -  & 55.2 & 25.2 & 21.3 & 25.4 & 50.4 & 35.5 \\
        \midrule

        \multirow{3}{*}{SliM-LLM~\cite{huang2024slim}} 
        & 4 & 16 &  128  & 78.9 & 79.9 & 49.4 & 58.7 & 72.6 & 67.9 \\
        & 3 & 16 &  128  & 77.8 & 73.7 & 42.9 & 55.5 & 72.8 & 64.5 \\
        & 2 & 16 &  128  & 57.1 & 35.4 & 26.1 & 28.9 & 56.6 & 40.8 \\
        \midrule
        
        \multirow{3}{*}{QuIP~\cite{chee2024quip}} 
        & 4 & 16 &  -  & 78.2 & 78.2 & 47.4 & 58.6 & 73.2 & 67.1 \\
        & 3 & 16 &  -  & 76.8 & 72.9 & 41.0 & 55.4 & 72.5 & 63.7 \\
        & 2 & 16 &  -  & 52.9 & 29.0 & 21.3 & 29.2 & 51.7 & 36.8 \\
        \midrule
        
        DB-LLM & 2 & 16 & 128 & 68.9 & 59.1 & 28.2 & 42.1 & 60.4 & 51.8 \\
        \midrule
        
        \multirow{2}{*}{PB-LLM~\cite{shang2023pb}}
        & 2 & 16 & 128 
        & 57.0 & 37.8 & 17.2 & 29.8 & 52.5 & 38.8\\
        &  1.7  &  16  &  128  
        & 52.5 & 31.7 & 17.5 & 27.7 & 50.4 & 36.0 \\
        \midrule
        
        BiLLM~\cite{huang2024billm} & 1.1 & 16 & 128 
        & 56.1 & 36.0 & 17.7 & 28.9 & 51.0 & 37.9\\
        \midrule
        \midrule

        \multirow{3}{*}{SmoothQuant~\cite{xiao2023smoothquant}} 
        & 8 & 8 & - & 79.5 & 79.7 & 49.0 & 60.0 & 73.2 & 68.3 \\
        & 6 & 6 & - & 76.8 & 75.5 & 45.0 & 56.9 & 69.0 & 64.6 \\
        & 4 & 4 & - & 54.6 & 26.3 & 20.0 & 26.4 & 50.3 & 35.5 \\ 
        \midrule

        \multirow{2}{*}{SpinQuant~\cite{liu2024spinquant}} 
        &  4 &  8 &  - &  79.6 &  76.5 &  54.0 &  78.1 &  72.4 &  72.1 \\
        &  4 &  4 &  - &  77.5 &  75.0 &  50.9 &  75.9 &  68.5 &  69.6 \\ 
        
        \bottomrule
        
    \end{tabular}}
\end{table*}

\begin{table*}[!t]
    \small
    \centering
    \caption{Evaluation results of post-training quantization on the LLaMA3-70B model (2/2). }

    \label{tab:ptq_lm370_2}
    \setlength{\tabcolsep}{3.8mm}
    {
    \begin{tabular}{lllrrrrrrrr}
        \toprule
        \multirow{2}{*}{{Method}} & 
        \multirow{2}{*}{{\#W }} & 
        \multirow{2}{*}{{\#A}} & 
        \multirow{2}{*}{{\#G}} & 
        \multicolumn{6}{c}{{CommonSenseQA$\uparrow$}} 
        \\
        \cmidrule(lr){5-10}
        ~ & ~ & ~ & ~  & {PIQA} &{ARC-e} & {ARC-c} &{HellaSwag} & {Wino} & {Avg.} \\

        \midrule
        LLaMA3 & 16 & 16 & - & 82.4 & 86.9 & 60.3 & 66.4 & 80.6 & 75.3 \\
        \midrule
        \midrule
        
        \multirow{3}{*}{RTN} 
        & 4 & 16 & 128
        & 82.3&85.2&58.4&65.6&79.8&74.3\\
        & 3 & 16 & 128 
        & 64.2 & 48.9 & 25.1& 41.1& 60.5&48.0\\
        & 2 & 16 & 128 
        & 53.2 & 23.9 & 22.1&25.8&53.0 & 35.6 \\
        \midrule
        
        \multirow{3}{*}{GPTQ~\cite{frantar2022gptq}} 
        & 4 & 16 & 128 
        &82.9&86.3&58.4&66.1&80.7&74.9 \\
        & 3 & 16 & 128 
        &80.6&79.6&52.1&63.5&77.1&70.6 \\
        & 2 & 16 & 128 
        &62.7&38.9&24.6&41.0&59.9&45.4  \\
        \midrule

        \multirow{3}{*}{AWQ~\cite{lin2023awq}} 
        & 4 & 16 & 128 & 82.7 & 86.3 & 59.0 & 65.7 & 80.9 & 74.9 \\
        & 3 & 16 & 128 & 81.4 & 84.7 & 58.0 & 63.5 & 78.6 & 73.2 \\
        & 2 & 16 & 128 & 52.2 & 25.5 & 23.1 & 25.6 & 52.3 & 35.7 \\
        \midrule

        \multirow{3}{*}{SliM-LLM~\cite{huang2024slim}} 
        & 4 & 16 &  128  & 82.9 & 86.5 & 59.0 & 66.2 & 80.7 & 75.1 \\
        & 3 & 16 &  128  &  81.6 & 83.1 & 58.5 & 64.7 & 78.4 & 73.3 \\
        & 2 & 16 &  128  &  76.2 & 66.3 & 45.7 & 55.4 & 63.7 & 61.5 \\ %
        \midrule
        
        \multirow{3}{*}{QuIP~\cite{chee2024quip}} 
        & 4 & 16 &  -  & 82.5 & 86.0 & 58.7 & 65.7 & 79.7 & 74.5 \\
        & 3 & 16 &  -  & 82.3 & 83.3 & 54.9 & 63.9 & 78.4 & 72.5 \\
        & 2 & 16 &  -  & 65.3 & 48.9 & 26.5 & 40.9 & 61.7 & 48.7 \\

        \midrule
        
        \multirow{2}{*}{PB-LLM~\cite{shang2023pb}}
        & 2 & 16 & 128 
        &65.2&40.6&25.1&42.7&56.4&46.0 \\
        & 1.7 & 16 & 128 
        &56.5&49.9&25.8&34.9&53.1&44.1\\
        \midrule
        
        BiLLM~\cite{huang2024billm} & 1.1 & 16 & 128 
        
        &58.2&46.4&25.1&37.5&53.6&44.2 \\ 
        \midrule
        \midrule

        \multirow{3}{*}{SmoothQuant~\cite{xiao2023smoothquant}} 
        & 8 & 8 & - & 82.2 & 86.9 & 60.2 & 66.3 & 80.7 & 75.3 \\
        & 6 & 6 & - & 82.4 & 87.0 & 59.9 & 66.1 & 80.6 & 75.2 \\
        & 4 & 4 & - & 76.9 & 75.8 & 43.5 & 52.9 & 58.9 & 61.6 \\   
        
        \bottomrule
        
    \end{tabular}}
\end{table*}

\begin{table*}[!t]
    \small
    \centering
    \caption{PPL results of post-training quantization on the LLaMA1/2-7B model}

    \label{tab:ptq_lm8_1}
    \setlength{\tabcolsep}{4.82mm}
    {
    \begin{tabular}{lllrrrrrrrrrrrrr}
        \toprule
        \multirow{2}{*}{{Method}} & 
        \multirow{2}{*}{{\#W }} & 
        \multirow{2}{*}{{\#A}} & 
        \multirow{2}{*}{{\#G}}&
        \multicolumn{2}{c}{{LLaMA-7B$\downarrow$}} &
        \multicolumn{2}{c}{{LLaMA2-7B$\downarrow$}}
        \\
        \cmidrule(lr){5-6} \cmidrule(lr){7-8}
        ~ & ~ & ~ & ~ & {WikiText2} & {C4} & {WikiText2} & {C4} \\ 
        \midrule
        FP & 16 & 16 & - & 5.7 & 7.1 & 5.5 & 7.0 \\
        \midrule
        \midrule
        
        \multirow{3}{*}{RTN} 
         & 4 & 16 & 128 & 6.0 & 7.4 & 5.7 & 7.2\\
         & 3 & 16 & 128 & 7.0 & 8.6 & 6.7 & 8.4 \\
         & 2 & 16 & 128 & 1.9\(\times 10^3\) & 1.0\(\times 10^3\) & 4.2\(\times 10^3\) & 4.9\(\times 10^3\) \\
        \midrule
        
        \multirow{3}{*}{GPTQ~\cite{frantar2022gptq}} 
         &  4  & 16 & 128 & 6.2 & - & 5.7 & - \\
         &  3  & 16 & 128 & 6.6 & 7.9 & 6.3 & 7.9 \\
         &  2  & 16 & 128 & 1.5\(\times 10^2\) & 34.6 & 60.5 & 33.7 \\
        \midrule

        \multirow{3}{*}{AWQ~\cite{lin2023awq}} 
        & 4 & 16 & 128 & 5.8 & - & 5.6 & - \\
        & 3 & 16 & 128 & 6.5 & 7.9 & 6.2 & 7.8 \\
        & 2 & 16 & 128 & 2.6\(\times 10^5\) & 1.9\(\times 10^5\) & 2.2\(\times 10^5\) & 1.75 \\
        \midrule

        \multirow{2}{*}{SliM-LLM~\cite{huang2024slim}} 
        & 3 & 16 & 128 & 6.4 & 6.1 & 6.2 & 7.7 \\
        & 2 & 16 & 128 & 14.6 & 32.9 & 16.0 & 16.0 \\
        \midrule
        
        \multirow{1}{*}{QuIP~\cite{chee2024quip}} 
        & 2 & 16 & - & 29.7 & 33.7 & 39.7 & 31.9 \\
        \midrule
        
        DB-LLM~\cite{chen2024db} & 2 & 16 & 128 & 7.6 & 9.7 & 7.2 & - \\
        \midrule
        
        \multirow{2}{*}{PB-LLM~\cite{shang2023pb}}
        & 2 & 16 & 128 & 24.6 & 49.7 & 25.4 & 29.8 \\
        & 1.7 & 16 & 128 & 1.0\(\times 10^2\) & 1.0\(\times 10^2\) & 69.2 & 80.2 \\
        \midrule
        
        BiLLM~\cite{huang2024billm} & 1.1 & 16 & 128 & 35.0 & 39.6 & 32.5 & 40.5 \\
        \midrule
        \midrule

        \multirow{2}{*}{SmoothQuant~\cite{xiao2023smoothquant}} 
        & 6 & 6 & - & 6.0 & 7.5  & 6.2 & 7.8 \\
        & 4 & 4 & - & 22.3 & 32.3 & 83.1 & 77.3 \\
        \midrule 
        
        \multirow{2}{*}{OmniQuant~\cite{shao2023omniquant}} 
        &  6 &  6 & - & 6.0 & 7.4 & 5.9 & 7.5 \\
        &  4 &  4 & - & 11.3 & 14.5 & 14.3 & 18.0 \\ 
        \midrule 
        
        \multirow{2}{*}{I-LLM~\cite{hu2024llm}} 
        &  6 &  6 & - & 5.8 & 7.3 & 5.7 & 7.3 \\
        &  4 &  4 & - & 9.1 & 12.3 & 10.4 & 12.9 \\ 
        \midrule 
        
        \multirow{2}{*}{SpinQuant~\cite{liu2024spinquant}} 
        &  4 &  8 & - & - & - & 5.7 & - \\
        &  4 &  4 & - & - & - & 5.9 & - \\ 
        
        \bottomrule
        
    \end{tabular}}
\end{table*}

\paragraph{PTQ methods\quad} 
{Among them, round-to-nearest (RTN) is a vanilla rounding quantization method that directly applies the statistical approach from Eq.\eqref{eq:quantization_ptq} to obtain quantization parameters for immediate quantization. GPTQ ~\cite{frantar2022gptq} is one of the most effective weight-only quantization methods, utilizing an error compensation strategy based on second-order loss. By using the inverse of the Hessian matrix, it reduces compression errors during quantization. AWQ ~\cite{lin2023awq} employs an activation-aware outlier suppression approach, introducing a scaling factor \( s \) to smooth the weight distribution of LLMs, thereby easing the quantization difficulty. QuIP ~\cite{chee2024quip} ensures consistency between weights and the Hessian by optimizing matrix computations and adopts codebook encoding to quantize weight parameters, further enhancing the mapping accuracy between continuous and discrete parameter spaces. Recently, Huang et al. ~\cite{huang2024slim} proposed a grouped mixed-precision quantization method that leverages the clustering characteristics of significant weights. This method uses mixed precision group quantization to achieve high-precision low-bit quantization in a hardware-friendly manner. Both approaches preserve LLaMA3’s 3-bit quantization capability, with the potential to bring 2-bit quantization to higher performance levels.}

The recent emergence of binarized LLM quantization methods has realized ultra-low bit-width LLM weight compression. PB-LLM~\cite{shang2023pb} employs a mixed-precision quantization strategy, retaining a small portion of significant weight full-precision while quantizing most of the weights to 1 bit. DB-LLM~\cite{chen2024db} achieves efficient LLM compression through double binarization weight splitting and proposes a deviation-aware distillation strategy to further improve 2-bit LLM performance. BiLLM~\cite{huang2024billm} pushes the LLM quantization limit further down to 1.1 bit by residual approximation of salient weights and grouped quantization of non-salient weights. These LLM quantization methods, which are specially designed for ultra-low bit-width, can achieve higher accuracy of quantized LLaMA3-8B at $\leqslant$ 2 bits, far outperforming methods such as GPTQ, AWQ, and QuIP below 2 bits (even 3 bits in some cases). We also perform the evaluation on quantized activations using  SmoothQuant~\cite{xiao2023smoothquant}, which shifts the quantization difficulty offline from activations to weights to smooth out activation outliers. Our evaluation shows that SmoothQuant can maintain the accuracy of LLaMA3 with 6/8-bit weights and activations, but collapses at 4 bits. Moreover, we find that the LLaMA3-70B model shows significant robustness to different quantization methods, even for ultra-low bit-width quantization.

{In the evaluation metrics of PPL (Table~\ref{tab:ptq_lm38_1} and Table~\ref{tab:ptq_lm370_1}) and CommonSenseQA (Table~\ref{tab:ptq_lm38_2} and Table~\ref{tab:ptq_lm370_2}), we found that, overall, the 4-bit methods had a slight performance decrease (approximately 2\%) compared to the original 16-bit LLM, with no significant differences between the different methods. In the context of 3-bit scenarios, traditional RTN quantization methods faced substantial performance losses (over 10\% lower than 4 bits), while methods such as GPTQ, AWQ, SliM-LLM, and QuIP were able to maintain performance close to that of 4 bits (with less than 5\% performance degradation). Interestingly, both DB-LLM and BiLLM were able to achieve reasonable results at ultra-low bit-width settings of 2 bits and even 1.1 bits, possibly due to the large-batch fine-tuning strategy and BiLLM's fine-grained salience partitioning. When quantifying both weight and activation simultaneously, both the 8B and 70B models demonstrated near lossless performance at 8 bits. As the bit-width was further reduced, the performance loss decreased significantly for the 8B model, while it decreased slowly for 70B models, indicating the presence of information redundancy within 70B models.}

{For practical deployment, we recorded the GPU memory usage and training time consumption for some PTQ methods on different sizes of the LLaMA model, as shown in Table~\ref{tab:deployment_ptq}. It demonstrates that methods such as SmoothQuant and AWQ are highly efficient in terms of memory usage and training time, with SmoothQuant requiring only 13.5 GB of GPU memory and 7 min for LLaMA2-7B, making it an ideal choice for memory-constrained environments. In contrast, OmniQuant, while effective for model compression, shows significantly higher quantization time consumption. Meanwhile, we tested the inference latency of the quantized 4-bit models resulting from the above methods in real-world deployment, as shown in Table~\ref{tab:deployment_ptq}. In fact, GPTQ, AWQ, and Omniquant all use block-wise quantization techniques, and theoretically, the upper bound of real inference speed optimization for these three methods is the same. To ensure a fair comparison of latency, we conducted tests using the deployment methods provided in the original methodology. In the case of LLaMA2-7B, GPTQ, AWQ, and Omniquant all exhibited speeds exceeding 100 tokens per second. However, in the case of LLaMA3-8B, the overall speed ranged between 50 to 80 tokens per second, with AWQ's quantization kernel achieving an inference speed of 89.8 tokens per second, surpassing that of other methods.}

\begin{table*}[!t]
    
    \small
    \centering
    \caption{GPU memory usage, quantization time, and inference latency for PTQ methods on LLaMA2-7B and LLaMA3-8B. Latency is determined under a group size of 128. '-' denotes that the current method did not provide the real quantization kernel for the latency test.}
    \label{tab:deployment_ptq}
    \setlength{\tabcolsep}{4.8mm}
    {
    \begin{tabular}{lcrrrrrr}
        \toprule
        \multirow{2}{*}{{Method}} & 
        \multirow{2}{*}{{\#W}} & 
        \multicolumn{3}{c}{{LLaMA2-7B}} & 
        \multicolumn{3}{c}{{LLaMA3-8B}} \\ 
        \cmidrule(lr){3-5} \cmidrule(lr){6-8}
        ~ & ~ & {\makecell{Memory \\ \footnotesize(GB)}} & {\makecell{Time \\ \footnotesize(min)}} &{\makecell{Speed \\ \footnotesize(token/s)}}& {\makecell{Memory \\ \footnotesize(GB)}}  & {\makecell{Time \\ \footnotesize(min)}}&{\makecell{Speed \\ \footnotesize(token/s)}} \\ 
        \midrule
        GPTQ~\cite{frantar2022gptq} & 4 & 26.4 & 17&159.4& 40.3 & 19&61.2\\
        SmoothQuant~\cite{xiao2023smoothquant} & 4 & 13.5 & 7&- & 16.0 & 15&- \\    
        AWQ~\cite{lin2023awq} & 4 & 11.7 & 12&112.9& 20.1 & 10&89.8\\    
        OmniQuant~\cite{shao2023omniquant} & 4 & 29.45 & 325&147.2& 30.61 & 307&54.9\\   
        \bottomrule
    \end{tabular}
    }
\end{table*}

\section{Track2: LoRA-FineTuning Quantization}\label{sec:qpeft} %

\paragraph{Quantization Framework\quad} 
{The LoRA-FT quantization process involves applying low-bit quantization to the original model weights, adding low-rank matrices to the pre-trained model weights, and fine-tuning the low-rank matrices with the training data, allowing model updates without modifying the core parameters. In addition to using the integer quantization commonly applied in PTQ, LoRA-FT can also use NormalFloat quantization. The NormalFloat quantization function for a weight matrix $\boldsymbol{w}_{q} \in \mathbb{R}^{n \times m} $ is defined as follows:}

{\begin{equation} \label{eq:quantization}\hat{\boldsymbol{w}}_q = \operatorname{NF}_k(\frac{\boldsymbol{w}}{s})\end{equation}}

{where $\hat{\boldsymbol{w}}_q$ indicates quantized weight, $s$ is the scale factor, typically set to the maximum value of $\boldsymbol{w}$ and $\operatorname{NF}_k$ denotes the NormalFloat quantization operator at $k$ bit-width, mapping each value in $\boldsymbol{w}_{\text{norm}}$ to the nearest quantile in the normal distribution for a bit-width $k$.}

\begin{table*}[t]
    \small
    \centering
    \caption{LoRA-FT on LLaMA3-8B with Alpaca dataset (1/2)}

    \label{tab:lora-ft_1}
    \setlength{\tabcolsep}{7.3mm}
    {
    \begin{tabular}{lcrrrrr}
        \toprule
        \multirow{2}{*}{{Method}} & 
        \multirow{2}{*}{{\#W}} & 
        \multicolumn{5}{c}{{MMLU$\uparrow$}}
        \\
        \cmidrule(lr){3-7} 
        ~ & ~ & {Hums.} & {STEM} & {Social} & {Other} & {Avg.}\\ 
        \midrule
        LLaMA3 & 16 & 59.0 & 55.3 & 76.0 & 71.5 & 64.8  \\
        NormalFloat & 4 & 56.8 & 52.9 & 73.6 & 69.4 & 62.5 \\    
        \midrule
        QLoRA~\cite{dettmers2024qlora} & 4 & 50.3 & 49.3 & 65.8 & 64.2 & 56.7  \\    
        IR-QLoRA~\cite{qin2024accurate} & 4 & 52.2 & 49.0 & 66.5 & 63.1 & 57.2 \\    
        \bottomrule
        
    \end{tabular}
    }
\end{table*}

\begin{table*}[!t]
    \small
    \centering
    \caption{LoRA-FT on LLaMA3-8B with Alpaca dataset (2/2)}

    \label{tab:lora-ft_2}
    \setlength{\tabcolsep}{5.4mm}
    {
    \begin{tabular}{lcrrrrrr}
        \toprule
        \multirow{2}{*}{{Method}} & 
        \multirow{2}{*}{{\#W}} & 
        \multicolumn{6}{c}{{CommonSenseQA$\uparrow$}} 
        \\
        \cmidrule(lr){3-8}
        ~ & ~ & {PIQA} &{ARC-e} & {ARC-c} &{HellaSwag} & {Wino} & {Avg.} \\ 
        \midrule
        LLaMA3 & 16 & 79.9 & 80.1 & 50.4 & 60.2 & 72.8 & 68.6 \\
        NormalFloat & 4 & 78.6 & 78.5 & 46.2 & 58.8 & 74.3 & 67.3 \\    
        \midrule
        QLoRA~\cite{dettmers2024qlora} & 4 & 76.6 & 74.8 & 45.0 & 59.4 & 67.0 & 64.5 \\    
        IR-QLoRA~\cite{qin2024accurate} & 4 & 76.3 & 74.3 & 45.3 & 59.1 & 69.5 & 64.9 \\    
        \bottomrule
        
    \end{tabular}
    }
\end{table*}

\begin{table*}[t]
    \small
     
    \centering
    \caption{LoRA-FT on LLaMA-7B with Alpaca dataset}

    \label{tab:lora-ft_1_lm1}
    \setlength{\tabcolsep}{7.3mm}
    {
    \begin{tabular}{lcrrrrr}
        \toprule
        \multirow{2}{*}{{Method}} & 
        \multirow{2}{*}{{\#W}} & 
        \multicolumn{5}{c}{{MMLU$\uparrow$}}
        \\
        \cmidrule(lr){3-7} 
        ~ & ~ & {Hums.} & {STEM} & {Social} & {Other} & {Avg.}\\ 
        \midrule
        LLaMA & 16 & 33.3 & 29.8 & 37.8 & 38.0 & 34.6 \\
        NormalFloat & 4 & 33.1 & 30.6 & 38.8 & 38.8 & 35.1 \\    
        \midrule
        QLoRA~\cite{dettmers2024qlora} & 4 & 36.1 & 31.9 & 42.0 & 44.5 & 38.4  \\    
        IR-QLoRA~\cite{qin2024accurate} & 4 & 38.6 & 34.6 & 45.2 & 45.5 & 40.8 \\    
        \bottomrule
    \end{tabular}
    }
\end{table*}

\paragraph{LoRA-FT Methods\quad} 
Except for the PTQ methods, we also provide the performance of 4-bit LLaMA3-8B with 2 different LoRA-FT quantization methods as shown in Table~\ref{tab:lora-ft_1} and Table~\ref{tab:lora-ft_2}, including QLoRA~\cite{dettmers2024qlora} and IR-QLoRA~\cite{qin2024accurate}. In addition, the performance of LLaMA-7B under the same setting is summarized in Table~\ref{tab:lora-ft_1_lm1}. {QLoRA~\cite{dettmers2024qlora} is the first LoRA-FT method that uses 4-bit NormalFloat quantization for base model weights, achieving significant memory reduction with minimal impact on model performance. Building on QLoRA, IR-QLoRA~\cite{qin2024accurate} introduces information calibration quantization and information elastic connection from the information inspection, resulting in high-performance adaptation with low-bit precision.}

On the MMLU dataset, the most notable observation with LLaMA3-8B under LoRA-FT quantization is that low-rank fine-tuning on the Alpaca~\cite{alpaca} dataset not only fails to compensate for the errors introduced by quantization, but actually exacerbates the degradation. Specifically, various LoRA-FT quantization methods yield worse performance for quantized LLaMA3 below 4 bits compared with their 4-bit counterparts without LoRA-FT. This is in stark contrast to similar phenomena on LLaMA and LLaMA2, where the 4-bit low-rank fine-tuned quantized versions for the front panel could even easily outperform the original FP16 counterpart on MMLU. According to our intuitive analysis, the main reason for this phenomenon is LLaMA3's strong performance due to its massive pre-scale training. This means that the performance loss due to the quantization of the original model cannot be compensated by fine-tuning on a tiny data set with low-rank parameters (which can be seen as a subset of the original model~\cite{hu2021lora,dettmers2024qlora}). 
Despite the significant quantization loss that cannot be compensated by fine-tuning, the 4-bit LoRA-FT quantized LLaMA3-8B significantly outperforms LLaMA-7B and LLaMA2-7B using different quantization methods. For instance, with the QLoRA method, the 4-bit LLaMA3-8B has an average accuracy of 57.0 (FP16: 64.8), exceeding the 4-bit  LLaMA-7B's 38.4 (FP16: 34.6) by 18.6, and surpassing the 4-bit  LLaMA2-7B's 43.9 (FP16: 45.5) by 13.1~\cite{xu2023qa,qin2024accurate}. 
This implies that a new LoRA-FT quantization paradigm is needed in the era of LLaMA3.

A similar phenomenon occurs with the CommonSenseQA benchmark. Compared to the 4-bit counterparts without LoRA-FT, the performance of the models fine-tuned using QLoRA and IR-QLoRA also declined ({e.g.} QLoRA 2.8\% vs. IR-QLoRA 2.4\% on average). This further demonstrates the strength of using high-quality datasets in LLaMA3, as the general dataset, Alpaca, does not contribute to the model's performance in other tasks. {Moreover, IR-QLoRA consistently outperforms QLoRA, due to its incorporation of information calibration quantization and information elastic connection through information inspection. These mechanisms allow IR-QLoRA to achieve high-performance adaptation even at low-bit accuracy.}

For practical deployment, we recorded the GPU memory usage and training time consumption for different sizes of the LLaMA model, as shown in Table~\ref{tab:deployment_lora_ft}. It demonstrates that both QLoRA and IR-QLoRA achieve significant memory efficiency, dramatically reducing the required memory footprint compared to the original LLaMA model. Nevertheless, both QLoRA and IR-QLoRA introduce inference bottlenecks primarily due to the dequantization process, which results in an increase in inference latency. The trade-off between the reduced memory footprint and the slight increase in latency is often acceptable for deployment in resource-constrained environments where memory is the limiting factor. Further optimizations, such as hardware-specific tuning and algorithmic improvements, could mitigate this bottleneck and improve overall inference speed.

\section{Track3: Multi-modal Large Language Model Quantization}\label{sec:mllmptq} 

For the MLLM model, we follow a common practice by conducting post-training quantization on the LLaMA3 part~\cite{lin2023awq,lin2024vila}. As shown in Table~\ref{tab:llava-next_1} and Table~\ref{tab:llava-next_2}, we compare the ultra-low bit-width performance of LLaVA-Next-8B under GPTQ and AWQ in six visual-language benchmarks.

We initially evaluate the pure language capabilities of LLaVA-Next-8B, as illustrated in Table~\ref{tab:llava-next_1}. The fp16 precision PPL metrics of the LLaMA3 model, after being fine-tuned for visual tasks, worsened across three datasets compared to its performance on language tasks. This also suggests that when fine-tuned for visual-language tasks, the introduction of image tokens leads to a partial loss and forgetting of LLaMA3’s inherent language abilities. The language capabilities of multi-modal LLMs (MLLMs) show a loss trend consistent with pure LLMs under low-bit quantization. Subsequently, we tested the quantized LLaMA3 within the MLLM model on visual QA tasks. As shown in Table~\ref{tab:llava-next_2}, under several advanced PTQ methods, the 4-bit MLLM exhibits a loss of less than 2\% on multi-modal benchmarks, efficiently performing visual-language tasks with reduced model size. 

\begin{table*}[!t]
    \small
    \centering
    \caption{GPU memory usage, training time, and inference latency for LoRA-FT Methods on LLaMA models. }
    \label{tab:deployment_lora_ft}
    \setlength{\tabcolsep}{4.3mm}
    {
    \begin{tabular}{lcrrrrrr}
        \toprule
        \multirow{2}{*}{{Method}} & 
        \multirow{2}{*}{{\#W}} & 
        \multicolumn{3}{c}{{LLaMA2-7B}} & 
        \multicolumn{3}{c}{{LLaMA3-8B}} \\ 
        \cmidrule(lr){3-5} \cmidrule(lr){6-8}
        ~ & ~ & {\makecell{Memory \\ \footnotesize(GB)}} & {\makecell{Time \\ \footnotesize(hour)}} &{\makecell{Speed \\ \footnotesize(token/s)}}& {\makecell{Memory \\ \footnotesize(GB)}}  & {\makecell{Time \\ \footnotesize(hour)}}&{\makecell{Speed \\ \footnotesize(token/s)}} \\ 
        \midrule
        LLaMA & 16 & - & - & 95.6 & - & - & 79.7 \\
        QLoRA~\cite{dettmers2024qlora}  & 4 & 7.2 & 15.3 & 88.6 & 13.2 & 16.1 & 72.8 \\
        IR-QLoRA~\cite{qin2024accurate} & 4 & 7.4 & 15.4 & 83.1 & 14.2 & 16.3 & 69.2 \\    
        \bottomrule
    \end{tabular}
    }
\end{table*}

\begin{table*}[!t]
    \small
    \centering
    \caption{Evaluation results of post-training quantization on LLaVA-Next-8B (1/2)}

    \label{tab:llava-next_1}
    \setlength{\tabcolsep}{8.35mm}
    {
    \begin{tabular}{lllrrr}
        \toprule
        \multirow{2}{*}{{Method}} & 
        \multirow{2}{*}{{\#W }} & 
        \multirow{2}{*}{{\#G}}&
        \multicolumn{3}{c}{{PPL$\downarrow$}}
        \\
        \cmidrule(lr){4-6} 
        ~ & ~ & ~ & {WikiText2} & {C4} & {PTB}\\ 
        \midrule
        \makecell{LLaVA-Next \\ (LLaMA3-8B)} & 16  & - & 9.5 & 14.8 & 16.3\\
        \midrule
        \midrule
                \multirow{3}{*}{RTN} 
         &  4   &  128  &  10.2  &  15.6  &  17.1 \\
         &  3  &  128  &  23.2  &  26.5  &  36.1  \\
         &  2   &  128  &  1.5\(\times 10^5\)  &  5.7\(\times 10^5\)  &  8.6\(\times 10^5\)  \\
        \midrule
        \multirow{3}{*}{GPTQ~\cite{frantar2022gptq}} 
         &  4   &  128  &  9.5  &  14.8  &  17.1 \\
         &  3  &  128  &  13.0  &  19.5  &  28.4  \\
         &  2   &  128  &  83.7  &  3.1\(\times 10^3\)  &  2.0\(\times 10^2\)  \\
        \midrule

        \multirow{3}{*}{AWQ~\cite{lin2023awq}} 
        & 4 & 128 & 9.9 & 15.3 & 16.9  \\
        & 3 & 128 & 11.7 & 17.9 & 20.2  \\
        & 2 & 128 & 1.6\(\times 10^6\) & 2.0\(\times 10^6\) & 2.2\(\times 10^6\) \\
  
        \midrule
        \multirow{3}{*}{SliM-LLM~\cite{huang2024slim}} 
        & 4  & 128 & 9.5 & 15.0 & 16.5 \\
        & 3 & 128 & 11.1 & 16.8 & 18.5 \\
        & 2 & 128 & 46.3 & 2.0\(\times 10^2\) & 1.8\(\times 10^2\)  \\ 
        
        \bottomrule
        
    \end{tabular}}
\end{table*}

\begin{table*}[!t]
    \small
    \centering
    \caption{Evaluation results of post-training quantization on LLaVA-Next-8B (2/2). N denots that the answer score is 0, or the outputs are unexpected characters.}

    \label{tab:llava-next_2}
    \setlength{\tabcolsep}{4.25mm}
    {
    \begin{tabular}{lllrrrrrr}
        \toprule
        \multirow{2}{*}{{Method}} & 
        \multirow{2}{*}{{\#W }} & 
        \multirow{2}{*}{{\#G}}&
        \multicolumn{6}{c}{{Multimodel Tasks$\uparrow$}} 
        \\
        \cmidrule(lr){4-9}
        ~ & ~ & ~  & {AI2D} &\makecell{{Chart}\\ {QA}} & \makecell{{Doc}\\ {VQA}} &\makecell{{MM}\\ {Bench}} & \makecell{{MME}\\ {Cognition}} & \makecell{{MME}\\ {Perception}}\\ 
        \midrule
        \makecell{LLaVA-Next \\ (LLaMA3-8B)} & 16  & - & 71.7 & 69.2 & 78.2 & 72.2 & 376.8 & 1588.3 \\
        \midrule
        \midrule

         \multirow{3}{*}{RTN} 
         &  4   &  128 
         & 70.4 & 65.7 & 77.0 & 70.1 & 304.4 & 1550.9  \\
         &  3  &  128 
         & 58.7 & 63.2 & 69.7 & 64.3 & 247.2 & 1526.2   \\
         &  2  &  128  
         &N&N&N&N&N&N \\
        \midrule
        
        \multirow{3}{*}{GPTQ~\cite{frantar2022gptq}} 
         &  4   &  128 
         & 70.7 & 67.4 & 77.4 & 71.0 & 331.6 & 1563.4  \\
         &  3  &  128 
         & 66.2 & 65.1 & 75.6 & 67.4 & 290.1 & 1541.7   \\
         &  2  &  128  
         &N&N&N&N&N&N \\
        \midrule

        \multirow{3}{*}{AWQ~\cite{lin2023awq}} 
        & 4 & 128 
        & 70.6 & 68.0 & 77.2 & 71.1 & 325.7 & 1562.7   \\
        & 3 & 128 
        & 67.7 & 65.4 & 74.4 & 68.0 & 298.6 & 1541.7   \\
        & 2 & 128 
        &N&N&N&N&N&N\\
  
        \midrule

        \multirow{3}{*}{SliM-LLM~\cite{huang2024slim}} 
        & 4 & 128 
        & 70.6 & 68.0 & 77.2 & 71.1 & 342.5 & 1563.9   \\
        & 3 & 128 
        & 68.2 & 67.5 & 74.8 & 68.9 & 321.0 & 1554.3   \\
        & 2 & 128 
        &57.2&49.3&60.6&60.9&282.1&1515.8\\
        
        \bottomrule
        
    \end{tabular}}
\end{table*}

At 3 bits, the performance loss ranges from 5\% to 20\%, with the highest loss, 20.75\%, occurring on the MME cognition task. Notably, regardless of GPTQ or AWQ, we observe that the 2-bit LLaVA-Next-8B completely collapses in the six multi-modal QA tasks, with scores dropping to zero. Although SliM-LLM mitigates the performance collapse of LLaVA-Next-8B at 2 bits, it still shows a large performance degradation.

\begin{figure}[!t]
  \centering
  \includegraphics[width=.47\textwidth]{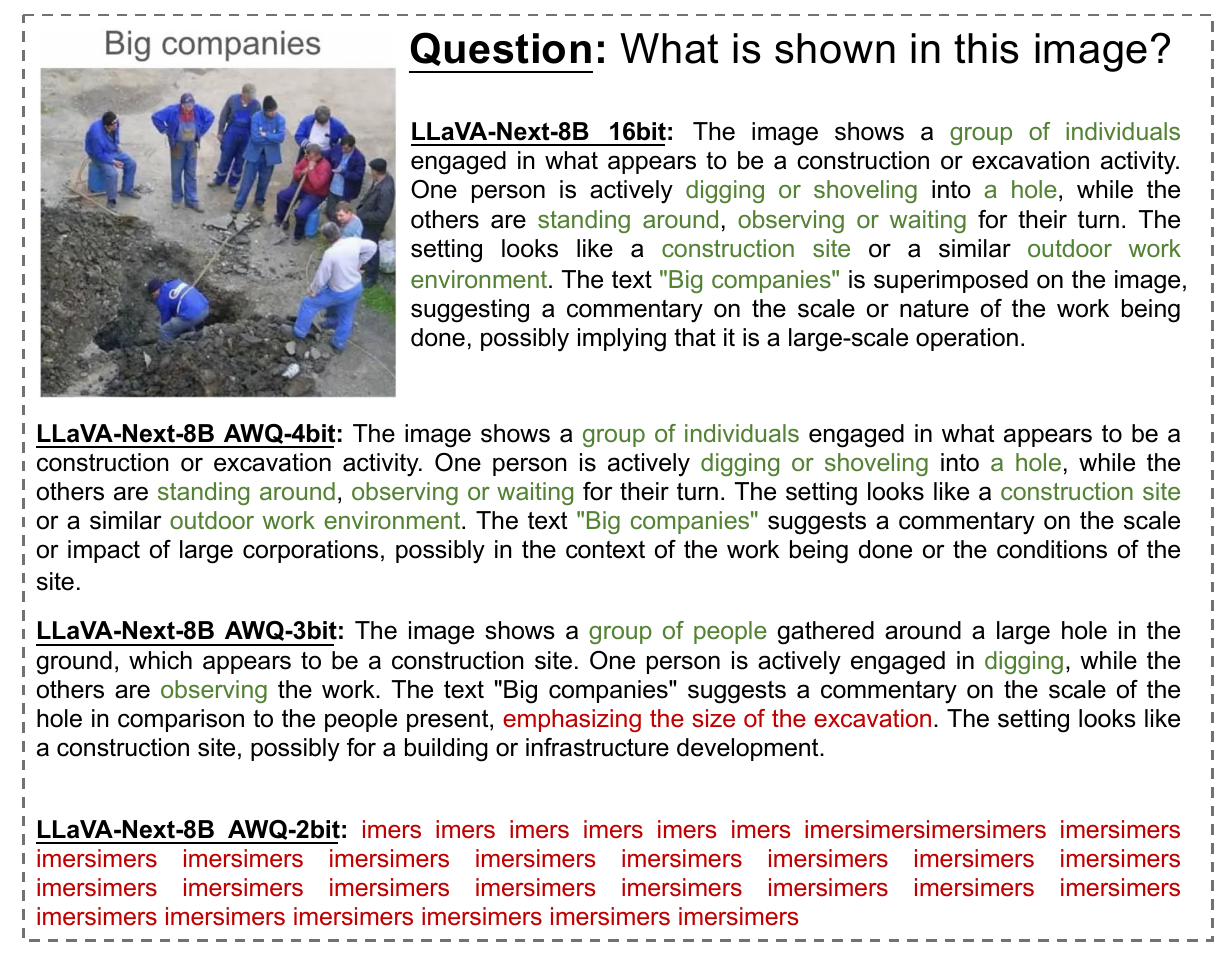}
  \caption{The VQA results of LLaVA-Next-8B for different quantization bit widths (1/5)}
  \label{fig:qa1}
\end{figure}

\begin{figure}[!t]
  \centering
  \includegraphics[width=.47\textwidth]{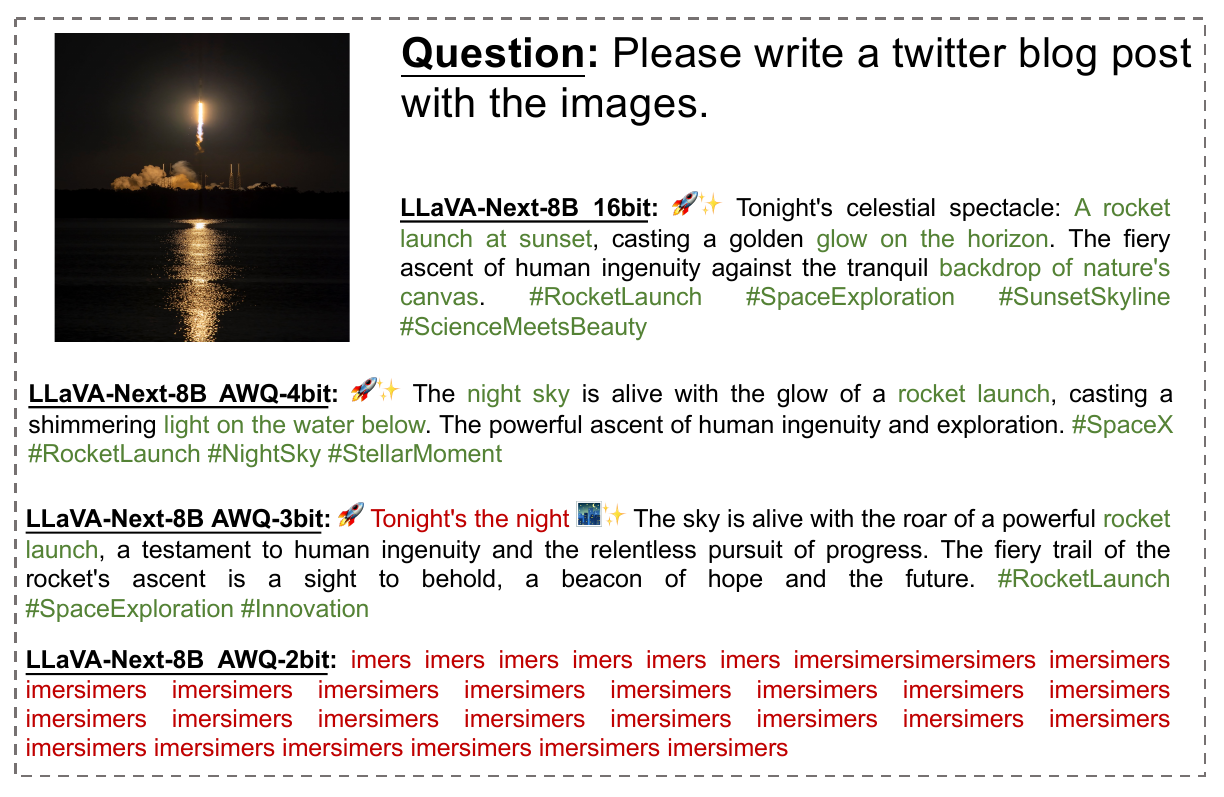}
  \caption{The VQA results of LLaVA-Next-8B for different quantization bit widths (2/5)}
  \label{fig:qa2}
\end{figure}

\begin{figure}[!t]
  \centering
  \includegraphics[width=.47\textwidth]{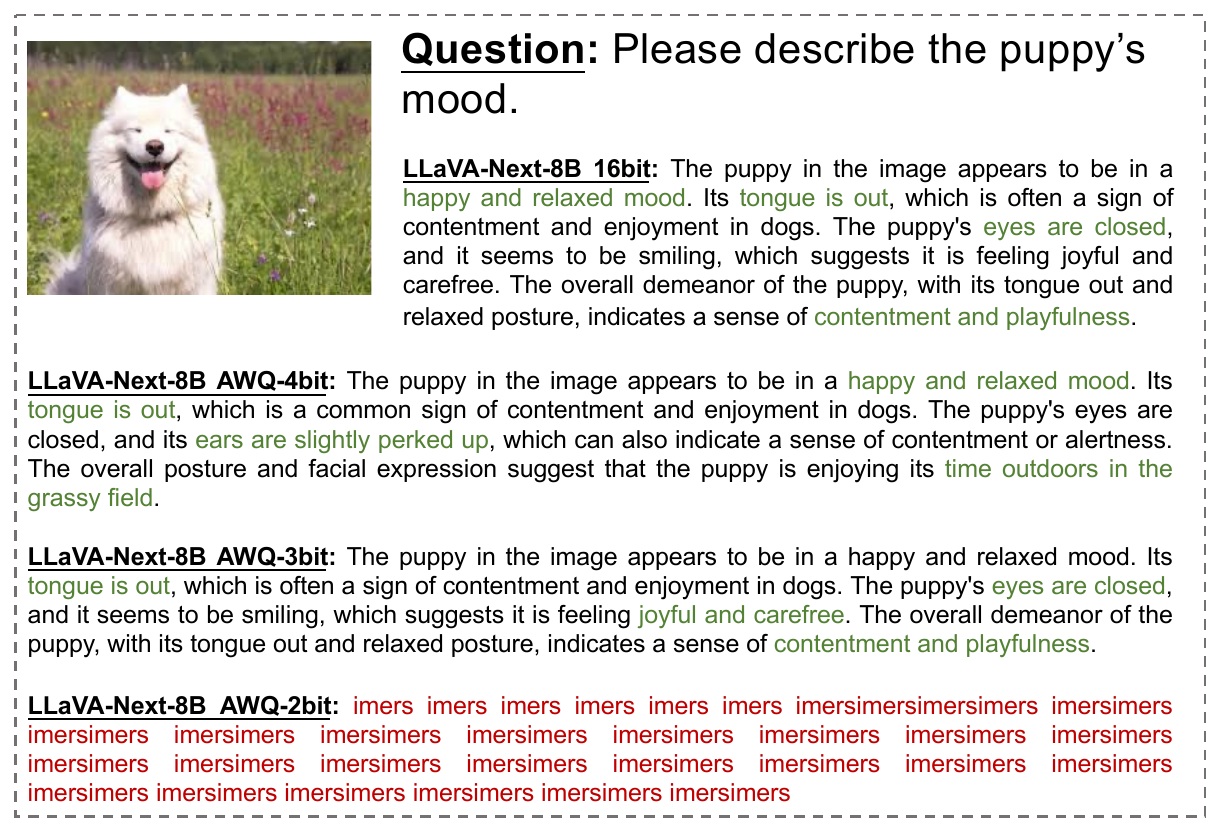}
  \caption{The VQA results of LLaVA-Next-8B for different quantization bit widths (3/5)}
  \label{fig:qa3}
\end{figure}

\begin{figure}[!t]
  \centering
  \includegraphics[width=.47\textwidth]{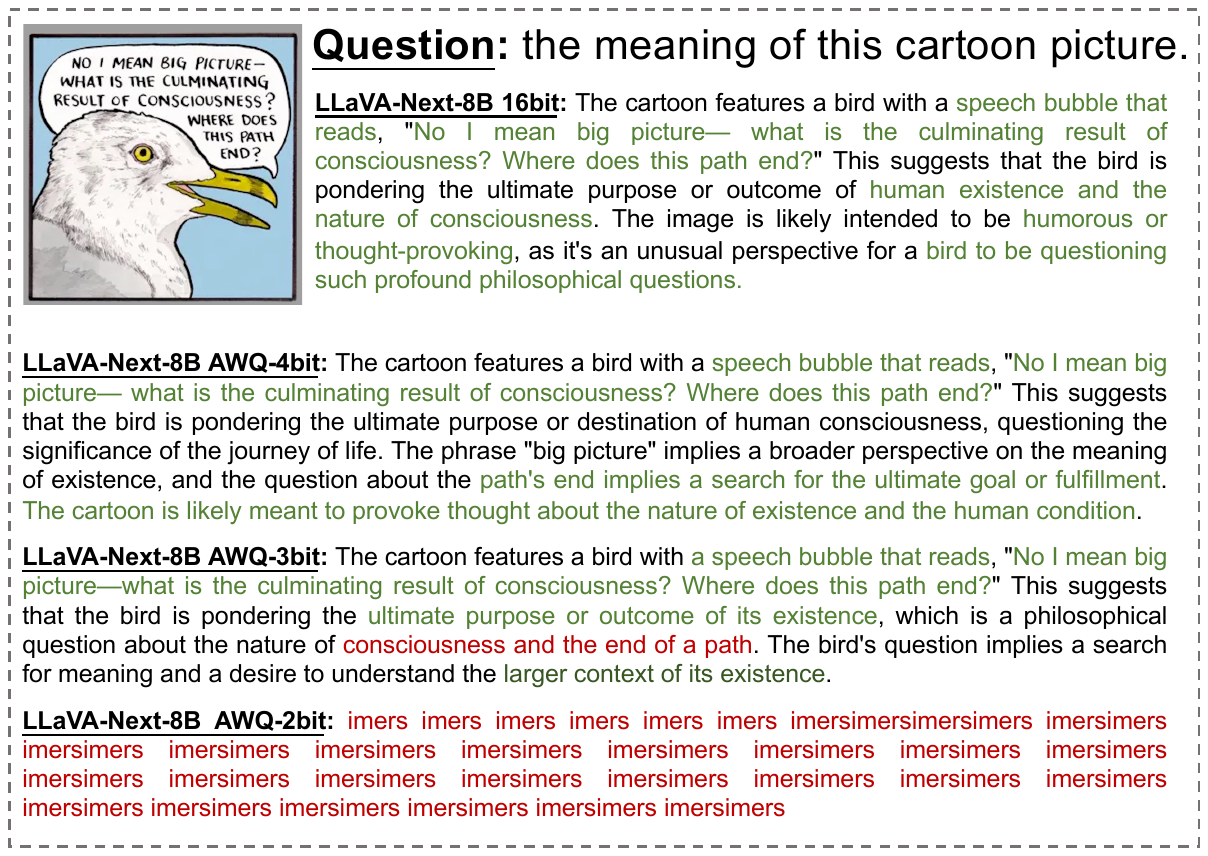}
  \caption{The VQA results of LLaVA-Next-8B for different quantization bit widths (4/5)}
  \label{fig:qa4}
\end{figure}

\begin{figure}[!t]
  \centering
  \includegraphics[width=.47\textwidth]{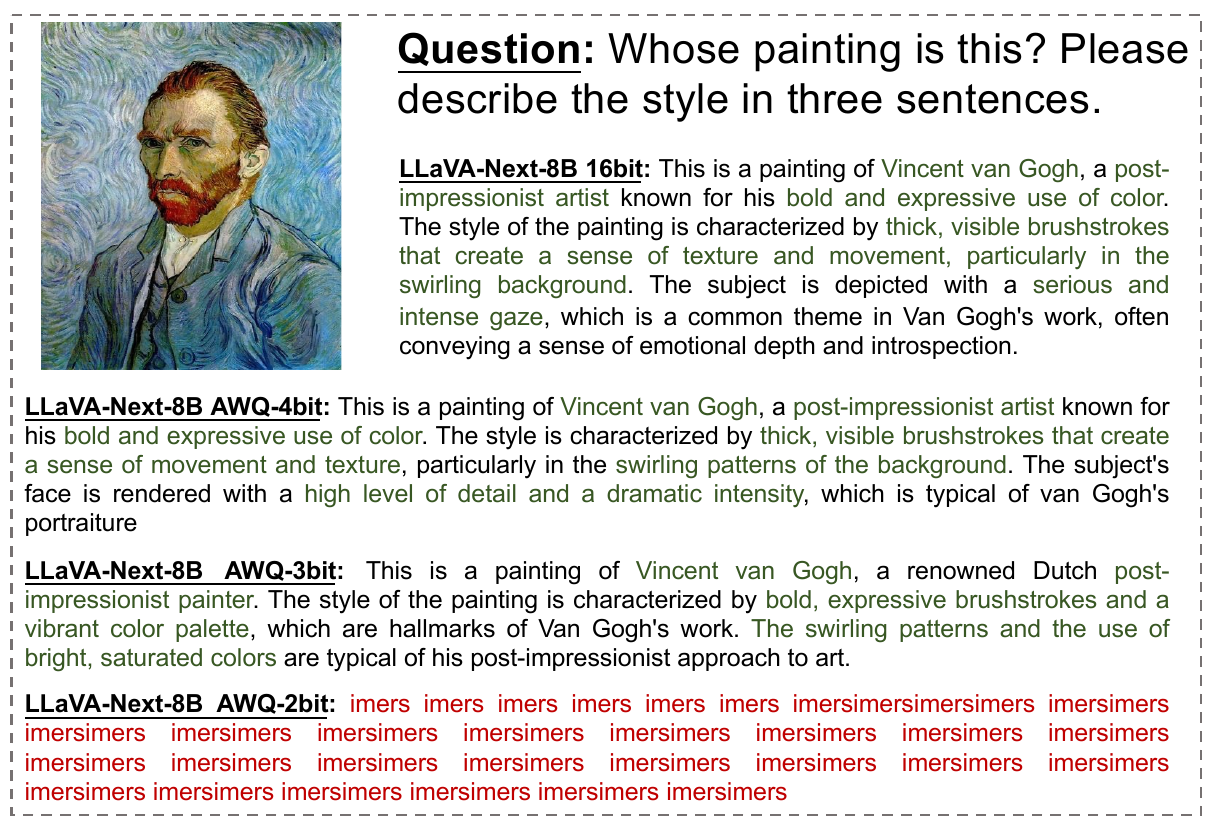}
  \caption{The VQA results of LLaVA-Next-8B for different quantization bit widths (5/5)}
  \label{fig:qa5}
\end{figure}

In Figs.~\ref{fig:qa1}-\ref{fig:qa5}, we show some real visual-language results of LLaVA-Next-8B under different bit widths quantized with AWQ. The 4-bit quantized model can still generate precise descriptions in images, while the 3-bit model excels in overall multi-modal understanding but suffers from a loss of detail. For example, in Fig.~\ref{fig:qa1}, the descriptions of people and actions in images by the 4-bit and 3-bit models are largely consistent with those of the 16-bit model. Additionally, the 4-bit model aligns with the 16-bit model in abstract semantic understanding of "big companies"; however, the 3-bit model misinterprets "big companies" as a descriptor of hole size. Further, under 2-bit quantization, the model struggles to produce reasonable answers, resulting in repetitive character responses. This contrasts with the performance of 2-bit models in pure language tasks, where previous studies~\cite{lin2023awq,huang2024billm,huang2024slim} have shown that 2-bit quantized models can still generate logically coherent sentences. However, in MLLM tasks, the 2-bit model fails to produce results close to expectations. This further indicates that the advanced PTQ method in the current LLM does not effectively perform equally well in the ultra-low bit MLLM models, which also inspires future work to propose better quantization solutions for this huge challenge in MLLM.



\section{Conclusion}\label{sec13}

The recently released LLaMA3 family has quickly become the most powerful LLM backbones, attracting significant interest from LLM and MLLM researchers. Building on this momentum, our study aims to thoroughly evaluate the performance of LLaMA3 for various low-bit quantization techniques, including post-training quantization and LoRA fine-tuning quantization for LLMs and MLLMs. Our goal is to assess the limits of its capabilities in resource-constrained scenarios using existing LLM and MLLM quantization techniques. 
We found that while LLaMA3 still demonstrates superior performance after quantization, the performance degradation associated with quantization is significant and can lead to larger declines. This decrease is mainly due to the fact that the powerful pre-training process allows LLaMA3 to learn more information to a similar extent as the previous LLaMA and LLaMA2, and its more sophisticated internal knowledge structure makes the effect of the quantization perturbation less obvious. The collapse of the ultra-low 2 bits also confirms that the quantized LLaMA3 backbone in MLLM exacerbates the performance loss caused by quantization when processing complex visual information.
This discovery highlights the potential challenges of deploying LLaMA3 in resource-constrained environments and underscores the ample room for growth and improvement in low-bit quantization. The empirical results of our research are expected to be valuable in the development of future LLM quantization techniques, especially in narrowing the performance gap with the original models. By addressing the performance degradation caused by low-bit quantization, we anticipate that subsequent quantization paradigms will allow LLMs to achieve stronger capabilities at a lower computational cost, ultimately driving the progress of generative artificial intelligence, as represented by LLMs and MLLMs, to new heights.

\vspace{.3in} \noindent \textbf{List of abbreviations:}
CV: computer vision; LLMs: large language models; LoRA-FT: LoRA-FineTuning; MLLMs: multi-modal large language model; NLU: natural language understanding; PTQ: post-training quantization

\begin{small}
\vspace{.3in} \noindent \textbf{Funding:}
This work was supported by the National Science and Technology Major Project (2021ZD0110503), the Swiss National Science Foundation (SNSF) project 200021E\_219943 Neuromorphic Attention Models for Event Data (NAMED), the Baidu Scholarship, and the National Natural Science Foundation of China (Nos. 62306025 and 92367204).

\vspace{.3in} \noindent \textbf{Data Availability:}
Availability of data and material: The datasets generated during and/or analyzed during the current study are available from the corresponding author on reasonable request. Our project is released on \href{https://github.com/Macaronlin/LLaMA3-Quantization}{GitHub}  and quantized LLaMA3 models are released in \href{https://huggingface.co/LLMQ}{HuggingFace}.

\vspace{.3in} \noindent \textbf{Competing Interests:}
All authors certify that they have no affiliations with or involvement in any organization or entity with any financial interest or non-financial interest in the subject matter or materials discussed in this manuscript.

\vspace{.3in} \noindent \textbf{Authors’ Contributions:}
All authors contributed to the study's conception and design. Wei Huang, Xingyu Zheng, Xudong Ma, Haotong Qin, Chengtao Lv, and Hong Chen performed data collection and analysis. Haotong Qin wrote the first draft of the manuscript, and all authors commented on previous versions. All authors read and approved the final manuscript. We propose the original idea together. 

\end{small}



\bibliographystyle{unsrt}
\bibliography{reference}

\begin{thebibliography}{10}

\bibitem{touvron2023llama}
Hugo Touvron, Thibaut Lavril, Gautier Izacard, Xavier Martinet, Marie-Anne Lachaux, Timoth{\'e}e Lacroix, et~al.
\newblock {LLaMA}: open and efficient foundation language models.
\newblock {\em arXiv preprint. arXiv:2302.13971}, 2023.

\bibitem{vaswani2017attention}
Ashish Vaswani, Noam Shazeer, Niki Parmar, Jakob Uszkoreit, Llion Jones, Aidan~N Gomez, et~al.
\newblock Attention is all you need.
\newblock In {\em I. Guyon, U. Von Luxburg, S. Bengio, {et al}. (Eds.), Proceedings of the 31st international conference on neural information processing systems}, pages 5998--6008. Red Hook: Curran Associates., 2017.

\bibitem{dubey2024llama}
Abhimanyu Dubey, Abhinav Jauhri, Abhinav Pandey, Abhishek Kadian, Ahmad Al-Dahle, Aiesha Letman, Akhil Mathur, Alan Schelten, Amy Yang, Angela Fan, et~al.
\newblock The llama 3 herd of models.
\newblock {\em arXiv preprint arXiv:2407.21783}, 2024.

\bibitem{liu2023llava}
Haotian Liu, Chunyuan Li, Qingyang Wu, and Yong~Jae \&~Lee.
\newblock Visual instruction tuning.
\newblock In {\em A. Oh, T. Neumann, A. Globerson, {et al}. (Eds.), Proceedings of the 37th international conference on neural information processing systems}, pages 1--25. Red Hook: Curran Associates., 2023.

\bibitem{xiao2023smoothquant}
Guangxuan Xiao, Ji~Lin, Mickael Seznec, Hao Wu, Julien Demouth, and Song \&~Han.
\newblock {SmoothQuant}: Accurate and efficient post-training quantization for large language models.
\newblock In {\em Proceedings of the International Conference on Machine Learning}, pages 38087--38099. Retrieved November 10, 2024, from \href{https://proceedings.mlr.press/v202/xiao23c.html}{https://proceedings.mlr.press/v202/xiao23c.html}., 2023.

\bibitem{qin2024quantsr}
Haotong Qin, Yulun Zhang, Yifu Ding, Xianglong Liu, Martin Danelljan, Fisher Yu, et~al.
\newblock Quant{SR}: accurate low-bit quantization for efficient image super-resolution.
\newblock In {\em A. Oh, T. Neumann, A. Globerson, {et al}. (Eds.), Proceedings of the 37th International Conference on neural information processing systems.}, pages 1--11. Red Hook: Curran Associates., 2023.

\bibitem{jacob2018quantization}
Benoit Jacob, Skirmantas Kligys, Bo~Chen, Menglong Zhu, Matthew Tang, Andrew Howard, et~al.
\newblock Quantization and training of neural networks for efficient integer-arithmetic-only inference.
\newblock In {\em Proceedings of the IEEE/CVF conference on computer vision and pattern recognition}, pages 2704--2713. Piscataway: IEEE., 2018.

\bibitem{huang2024slim}
Wei Huang, Haotong Qin, Yangdong Liu, Yawei Li, Xianglong Liu, Luca Benini, et~al.
\newblock {SliM-LLM}: Salience-driven mixed-precision quantization for large language models.
\newblock {\em arXiv preprint. arXiv:2405.14917}, 2024.

\bibitem{nagel2020up}
Markus Nagel, Rana~Ali Amjad, Mart Van~Baalen, Christos Louizos, and Tijmen Blankevoort.
\newblock Up or down? adaptive rounding for post-training quantization.
\newblock In {\em International Conference on Machine Learning}, pages 7197--7206. PMLR, 2020.

\bibitem{frantar2022gptq}
Elias Frantar, Saleh Ashkboos, Torsten Hoefler, and Dan \&~Alistarh.
\newblock {GPTQ}: Accurate post-training quantization for generative pre-trained transformers.
\newblock {\em arXiv preprint. arXiv:2210.17323}, 2022.

\bibitem{lin2023awq}
Ji~Lin, Jiaming Tang, Haotian Tang, Shang Yang, Wei-Ming Chen, Guangxuan Xiao, et~al.
\newblock {AWQ}: Activation-aware weight quantization for on-device {LLM} compression and acceleration.
\newblock In {\em P. B. Gibbons, G. Pekhimenko, \& C. de Sa (Eds.), Proceedings of Machine Learning and Systems}, pages 87--100. Retrieved November 10, 2024, from \href{https://proceedings.mlsys.org/paper\_files/paper/2024/hash/42a452cbafa9dd64e9ba4aa95cc1ef21-Abstract-Conference.html}{https://proceedings.mlsys.org/paper\_files/paper/2024/hash/42a452cbafa9dd64e9ba4aa95cc1ef21-Abstract-Conference.html}., 2024.

\bibitem{shang2023pb}
Yuzhang Shang, Zhihang Yuan, Qiang Wu, and Zhen \&~Dong.
\newblock {PB-LLM}: Partially binarized large language models.
\newblock In {\em Proceedings of the 12th international conference on learning representations}, pages 1--14. Retrieved November 10, 2024, from \href{https://openreview.net/forum?id=BifeBRhikU}{https://openreview.net/forum?id=BifeBRhikU}., 2024.

\bibitem{chee2024quip}
Jerry Chee, Yaohui Cai, Volodymyr Kuleshov, and Christopher \&~De~Sa.
\newblock {QuIP}: 2-bit quantization of large language models with guarantees.
\newblock In {\em A. Oh, T. Neumann, A. Globerson, {et al}. (Eds.), Proceedings of the 37th international conference on neural information processing systems}, pages 1--34. Red Hook: Curran Associates., 2024.

\bibitem{chen2024db}
Hong Chen, Chengtao Lv, Liang Ding, Haotong Qin, Xiabin Zhou, Yifu Ding, et~al.
\newblock {DB-LLM}: Accurate dual-binarization for efficient {LLMs}.
\newblock In {\em L.-W. Ku, A. Martins, V. Srikumar {(Eds.)}, Findings of the Association for Computational Linguistics}, pages 8719--8730. Stroudsburg: ACL.

\bibitem{huang2024billm}
Wei Huang, Yangdong Liu, Haotong Qin, Ying Li, Shiming Zhang, Xianglong Liu, et~al.
\newblock {BiLLM}: Pushing the limit of post-training quantization for {LLMs}.
\newblock In {\em Proceedings of the 41st international conference on machine learning}, pages 1--20. Retrieved November 10, 2024, from \href{https://openreview.net/forum?id=qOl2WWOqFg}{https://openreview.net/forum?id=qOl2WWOqFg}., 2024.

\bibitem{dettmers2024qlora}
Tim Dettmers, Artidoro Pagnoni, Ari Holtzman, and Luke \&~Zettlemoyer.
\newblock {QLoRA}: Efficient finetuning of quantized {LLMs}.
\newblock In {\em A. Oh, T. Neumann, A. Globerson, {et {al}}. (Eds.), Proceedings of the 37th international conference on neural information processing systems}, pages 1--28. Red Hook: Curran Associates., 2024.

\bibitem{qin2024accurate}
Haotong Qin, Xudong Ma, Xingyu Zheng, Xiaoyang Li, Yang Zhang, Shouda Liu, et~al.
\newblock Accurate lora-finetuning quantization of {LLMs} via information retention.
\newblock In {\em Proceedings of the 41st international conference on machine learning}, pages 1--19. Retrieved November 10, 2024, from \href{https://openreview.net/forum?id=jQ92egz5Ym}{https://openreview.net/forum?id=jQ92egz5Ym}., 2024.

\bibitem{merity2016pointer}
Stephen Merity, Caiming Xiong, James Bradbury, and Richard \&~Socher.
\newblock Pointer sentinel mixture models.
\newblock In {\em Proceedings of the 5th International Conference on Learning Representations}, pages 1--15. Retrieved November 10, 2024, from \href{https://openreview.net/forum?id=Byj72udxe}{https://openreview.net/forum?id=Byj72udxe}., 2016.

\bibitem{raffel2020exploring}
Colin Raffel, Noam Shazeer, Adam Roberts, Katherine Lee, Sharan Narang, Michael Matena, et~al.
\newblock Exploring the limits of transfer learning with a unified text-to-text transformer.
\newblock {\em The Journal of Machine Learning Research}, 21(1):5485--5551, 2020.

\bibitem{marcus1994penn}
Mitch Marcus, P.Grace Kim, Mary~Ann Marcinkiewicz, Robert MacIntyre, Ann Bies, Ferguson M., et~al.
\newblock The {Penn Treebank}: Annotating predicate argument structure.
\newblock In {\em Proceedings of Human Language Technology workshop}, pages 114--119. San Francisco: Morgan Kaufmann, 1994.

\bibitem{bisk2020piqa}
Y~Bisk, R~Zellers, R~Le~Bras, J~Gao, and Y~\&~Choi.
\newblock {PIQA}: Reasoning about physical commonsense in natural language.
\newblock In {\em Proceedings of the 34th AAAI conference on artificial intelligence}, pages 7432--7439. Palo Alto: AAAI Press., 2020.

\bibitem{clark2018think}
Peter Clark, Isaac Cowhey, Oren Etzioni, Tushar Khot, Ashish Sabharwal, Carissa Schoenick, et~al.
\newblock Think you have solved question answering? {Try ARC-DA}, the {AI2} reasoning challenge.
\newblock {\em arXiv preprint. arXiv:1803.05457}, 2018.

\bibitem{zellers2019hellaswag}
Rowan Zellers, Ari Holtzman, Yonatan Bisk, Ali Farhadi, and Yejin \&~Choi.
\newblock {HellaSwag}: Can a machine really finish your sentence?
\newblock In {\em A. Korhonen, D. R. Traum, L. M`arquez {(Eds.)}, Proceedings of the 57th conference of the Association for Computational Linguistics}, pages 4791--4800. Stroudsburg: ACL, 2019.

\bibitem{sakaguchi2021winogrande}
Keisuke Sakaguchi, Ronan~Le Bras, Chandra Bhagavatula, and Yejin \&~Choi.
\newblock Winogrande: An adversarial winograd schema challenge at scale.
\newblock {\em Communications of the ACM}, 64(9):99--106, 2021.

\bibitem{hendryckstest2021}
Dan Hendrycks, Collin Burns, Steven Basart, Andy Zou, Mantas Mazeika, Dawn Song, and Jacob Steinhardt.
\newblock Measuring massive multitask language understanding.
\newblock {\em Proceedings of the International Conference on Learning Representations (ICLR)}, 2021.

\bibitem{kembhavi2016diagram}
Aniruddha Kembhavi, Mike Salvato, Eric Kolve, Minjoon Seo, Hannaneh Hajishirzi, and Ali Farhadi.
\newblock A diagram is worth a dozen images, 2016.

\bibitem{masry2022chartqa}
Ahmed Masry, Do~Xuan Long, Jia~Qing Tan, Shafiq Joty, and Enamul Hoque.
\newblock Chartqa: A benchmark for question answering about charts with visual and logical reasoning.
\newblock {\em arXiv preprint arXiv:2203.10244}, 2022.

\bibitem{mathew2021docvqa}
Minesh Mathew, Dimosthenis Karatzas, and CV~Jawahar.
\newblock Docvqa: A dataset for vqa on document images.
\newblock In {\em Proceedings of the IEEE/CVF winter conference on applications of computer vision}, pages 2200--2209, 2021.

\bibitem{fu2024mmecomprehensiveevaluationbenchmark}
Chaoyou Fu, Peixian Chen, Yunhang Shen, Yulei Qin, Mengdan Zhang, Xu~Lin, Jinrui Yang, Xiawu Zheng, Ke~Li, Xing Sun, Yunsheng Wu, and Rongrong Ji.
\newblock Mme: A comprehensive evaluation benchmark for multimodal large language models, 2024.

\bibitem{liu2025mmbench}
Yuan Liu, Haodong Duan, Yuanhan Zhang, Bo~Li, Songyang Zhang, Wangbo Zhao, Yike Yuan, Jiaqi Wang, Conghui He, Ziwei Liu, et~al.
\newblock Mmbench: Is your multi-modal model an all-around player?
\newblock In {\em European Conference on Computer Vision}, pages 216--233. Springer, 2025.

\bibitem{hendrycks2020measuring}
Dan Hendrycks, Collin Burns, Steven Basart, Andy Zou, Mantas Mazeika, Dawn Song, et~al.
\newblock Measuring massive multitask language understanding.
\newblock In {\em Proceedings of the 9th international conference on learning representations}, pages 1--27. Retrieved November 10, 2024, from \href{https://openreview.net/forum?id=d7KBjmI3GmQ}{https://openreview.net/forum?id=d7KBjmI3GmQ}., 2021.

\bibitem{liu2023llm}
Zechun Liu, Barlas Oguz, Changsheng Zhao, Ernie Chang, Pierre Stock, Yashar Mehdad, et~al.
\newblock {LLM-QAT}: Data-free quantization aware training for large language models.
\newblock In {\em L.-W. Ku, A. Martins, V. Srikumar (Eds.), Findings of the Association for Computational Linguistics}, pages 467--484. Stroudsburg: ACL, 2024.

\bibitem{shao2023omniquant}
Wenqi Shao, Mengzhao Chen, Zhaoyang Zhang, Peng Xu, Lirui Zhao, Zhiqian Li, Kaipeng Zhang, Peng Gao, Yu~Qiao, and Ping Luo.
\newblock Omniquant: Omnidirectionally calibrated quantization for large language models.
\newblock {\em arXiv preprint arXiv:2308.13137}, 2023.

\bibitem{hu2024llm}
Xing Hu, Yuan Cheng, Dawei Yang, Zhihang Yuan, Jiangyong Yu, Chen Xu, and Sifan Zhou.
\newblock I-llm: Efficient integer-only inference for fully-quantized low-bit large language models.
\newblock {\em arXiv preprint arXiv:2405.17849}, 2024.

\bibitem{liu2024spinquant}
Zechun Liu, Changsheng Zhao, Igor Fedorov, Bilge Soran, Dhruv Choudhary, Raghuraman Krishnamoorthi, Vikas Chandra, Yuandong Tian, and Tijmen Blankevoort.
\newblock Spinquant--llm quantization with learned rotations.
\newblock {\em arXiv preprint arXiv:2405.16406}, 2024.

\bibitem{alpaca}
Rohan Taori, Ishaan Gulrajani, Tianyi Zhang, Yann Dubois, Xuechen Li, Carlos Guestrin, et~al.
\newblock Stanford alpaca: An instruction-following llama model.
\newblock Retrieved November 10, 2024, from \url{https://github.com/tatsu-lab/stanford_alpaca}, 2023.

\bibitem{hu2021lora}
Edward~J Hu, Phillip Wallis, Zeyuan Allen-Zhu, Yuanzhi Li, Shean Wang, Lu~Wang, Weizhu Chen, et~al.
\newblock {LoRA}: Low-rank adaptation of large language models.
\newblock In {\em Proceedings of the 10th International Conference on Learning Representations}, pages 1--13. Retrieved November 10, 2024, from \href{https://openreview.net/forum?id=nZeVKeeFYf9}{https://openreview.net/forum?id=nZeVKeeFYf9}., 2022.

\bibitem{xu2023qa}
Yuhui Xu, Lingxi Xie, Xiaotao Gu, Xin Chen, Heng Chang, Hengheng Zhang, et~al.
\newblock {QA-LoRA}: Quantization-aware low-rank adaptation of large language models.
\newblock In {\em Proceedings of the 12th International Conference on Learning Representations}, pages 1--18. Retrieved November 10, 2024, from \href{https://openreview.net/forum?id=WvFoJccpo8}{https://openreview.net/forum?id=WvFoJccpo8}., 2024.

\bibitem{lin2024vila}
Ji~Lin, Hongxu Yin, Wei Ping, Pavlo Molchanov, Mohammad Shoeybi, and Song \&~Han.
\newblock {VILA}: On pre-training for visual language models.
\newblock In {\em Proceedings of the IEEE/CVF Conference on Computer Vision and Pattern Recognition}, pages 26689--26699. Piscataway: IEEE., 2024.

\end{thebibliography}

\end{document}